\documentclass[letterpaper,10pt,journal,twoside]{IEEEtran}
\IEEEoverridecommandlockouts

\usepackage{amsfonts,amssymb,amsmath,mathtools,multicol}
\usepackage{physics}

\usepackage{float}
\usepackage{graphicx}
\graphicspath{{figures/out/}}
\usepackage{subfig}
\usepackage[font=footnotesize]{caption}
\usepackage[bibstyle=ieee,citestyle=numeric-comp,sortcites,mincitenames=1,maxcitenames=2,natbib=true,backend=biber]{biblatex}
\AtBeginBibliography{\footnotesize}
\usepackage[hidelinks]{hyperref}

\usepackage{indentfirst}

\usepackage{enumitem}

\usepackage[dvipsnames,table,xcdraw]{xcolor}
\usepackage{balance}

\newcommand{\R}{\mathbb{R}} 

\newcommand{\V}[1]{\boldsymbol{\mathbf{#1}}} 
\newcommand{\T}{\top}

\title{Keeping Contact: Task-Adaptive Friction Margins for Bimanual Non-Prehensile Manipulation}
\author{Qian Ying Lee$^1$, Lin Yang$^1$, and Domenico Campolo$^{1,2,\ast}$%
\thanks{$^{1}$School of Mechanical and Aerospace Engineering, Nanyang Technological University (NTU), Singapore.}%
\thanks{$^{2}$R3 Institute (R3I) Singapore, Nanyang Technological University (NTU), Singapore.}%
\thanks{$^{\ast}$Corresponding author: {\tt\footnotesize d.campolo@ntu.edu.sg}}
}

\begin{document}
\maketitle

\begin{abstract}
    Maintaining contact in non-prehensile manipulation requires more than nominal friction cone feasibility: a force barely inside the cone and one well inside it are both feasible, but they provide different margins against slip and contact loss. This paper uses signed friction distance to measure friction margin in force space and derives a task-adaptive margin from the force correction needed to balance pivot moment deviations in bimanual tilting. The resulting contact margins are embedded in a sequence of second-order cone programs (SOCPs), each giving a globally optimal contact force solution at a tilt angle. The optimizer balances the object about an environmental pivot while keeping robot and pivot contact forces away from the nominal friction boundary. Optimizer-level results show interpretable force redistribution and adaptation to object property changes. Real-world experiments demonstrate successful execution under object property variations and model mismatch, while fixed margin and no-margin comparisons illustrate the importance of adapting the margin to task demand. These results support signed friction distance margins as a force planning mechanism for maintaining contact in bimanual non-prehensile manipulation. Project page: \url{https://keeping-contact.github.io/}
\end{abstract}

\section{Introduction}\label{sec:introduction}
Non-prehensile manipulation extends robotic manipulation beyond rigid grasping by exploiting contact interactions as part of the task mechanics. This enables behaviors such as pushing \cite{heins2024force}, rolling \cite{woodruff2023robotic}, and pivoting \cite{zhang2021manipulation}. Since the resulting motion is governed by frictional contacts, planning must reason about both contact motion and forces. Recent optimization-based formulations incorporate trajectory optimization with frictional contact constraints and simultaneous contact selection \cite{shirai2025hierarchical,zhang2025simultaneous}. In these formulations, friction constraints primarily specify admissible contact forces or contact modes along the motion. Yet admissibility does not specify where a maintained contact force lies within the friction cone: a nominally feasible force profile may still operate near a friction boundary and remain sensitive to modeling error or tracking deviation.

A useful clue comes from everyday manipulation. When lifting an unfamiliar object, a person rarely knows its inertial parameters exactly, yet contact is often maintained stably. Rather than regulating contact force at the minimum value that barely avoids slip, people maintain a margin above the slip threshold and adjust it as perceived task demand changes \cite{westling1984factors, flanagan1997role}. This contrasts with friction cones used only as feasibility tests: a force that barely clears the boundary and one well inside the cone are accepted alike, although only the latter carries a meaningful margin. The question is thus not merely whether a planned force is feasible, but what margin it should maintain and how that margin should vary with the task.

\begin{figure}[t]
    \centering
    \includegraphics[width=\linewidth]{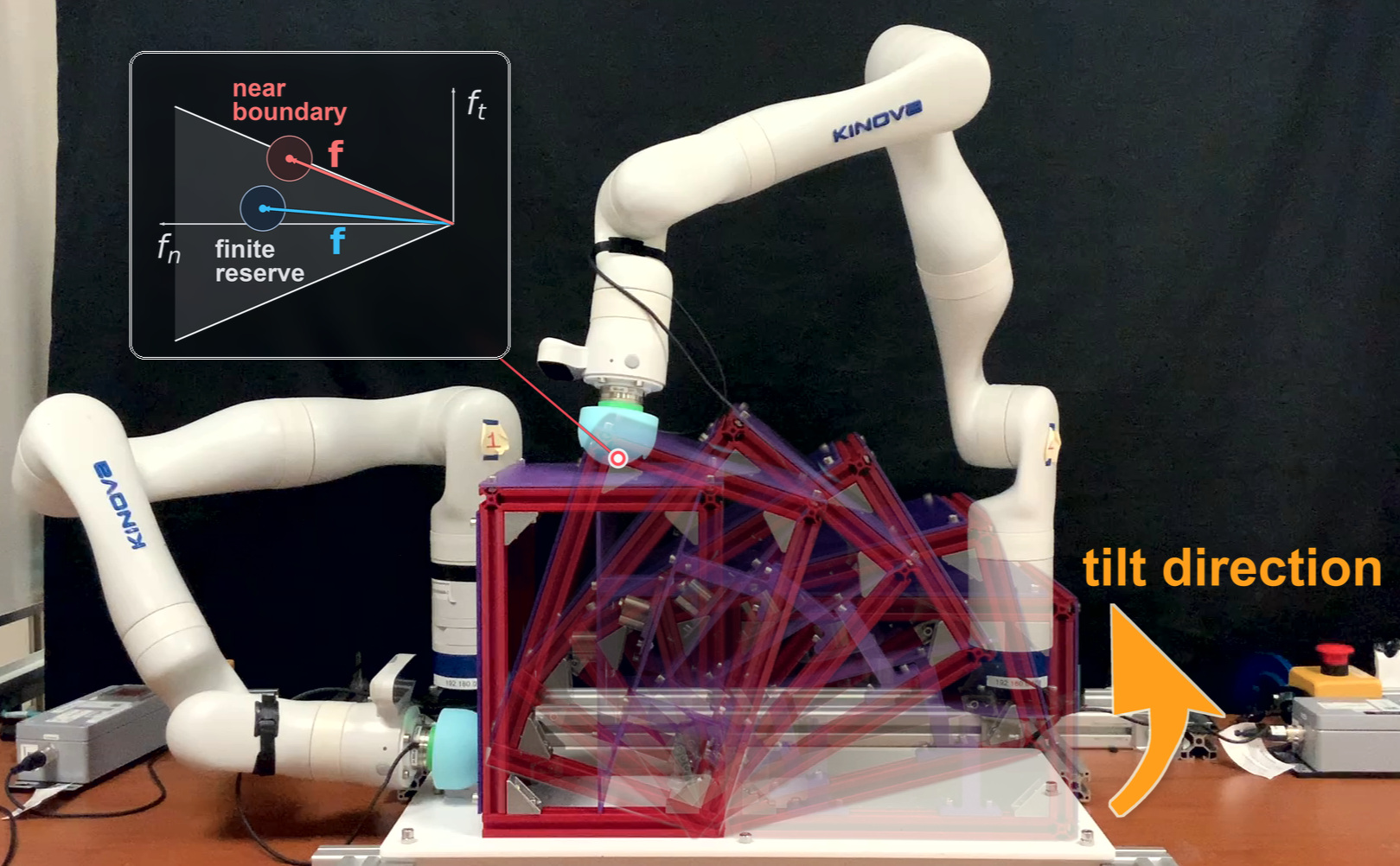}
    \caption{Bimanual non-prehensile tilting. Inset illustrates two feasible contact forces in the local force plane. Translucent circles show equal force perturbations around each force tip: a near boundary force can leave the friction cone and slip, whereas a force with finite margin remains feasible.}
    \label{fig:overview}
\end{figure}

Motivated by this view, we plan contact forces by lower bounding their signed distance from the nominal friction boundary. This lower bound gives each maintained contact a force space margin against deviations from the planned force. The remaining question is how to choose this margin along the motion: a fixed value treats margin as uniform, while a task-adaptive value can vary with the force demand imposed by the manipulation task. We study this construction on the planar bimanual tilting task in Fig.~\ref{fig:overview}. This task has a structured but nonuniform force requirement: the gravitational moment varies with tilt angle and object properties, changing the forces required from the two robot contacts. The motion is especially sensitive to slip at the initial and final configurations, making sufficient tangential support and friction margin important throughout the tilt. Unlike prior pivoting studies that often rely on symmetric robot--object contact arrangements \cite{fakhari2021motion,zhang2021manipulation}, our setup uses non-opposing contacts and exploits an environmental pivot to accomplish the task. Maintaining both robot contacts is therefore central to the task, since slip or contact loss at either contact can lead to task failure.

In summary, the main contributions of this work are:
\begin{enumerate}[leftmargin=*]
    \item We use signed friction distance to the nominal friction boundary as an explicit force space measure of margin for contact force planning.
    \item We introduce a task-adaptive margin construction that lower bounds this signed distance based on the force demand required by the task.
    \item We embed this margin in a convex force planner with globally optimal contact force solutions and evaluate it in real-world experiments.
\end{enumerate}

\section{Related Work}\label{sec:related-work}

\subsection{Non-Prehensile Manipulation with Force Planning}
Prior work in non-prehensile manipulation has shown that useful object motion can be generated by exploiting frictional and environmental contacts instead of rigidly grasping the object \cite{ruggiero2018nonprehensile,aiyama1993pivoting,dafle2014extrinsic}. Existing methods have studied contact mode planning, reorientation with environmental support, and bimanual coordination under changing contact conditions \cite{hou2020manipulation,cheng2021contact,wang2022contact,fakhari2021motion,zhang2021manipulation,dileva2024force}. These works demonstrate the value of planning through contact, but many emphasize object motion, contact modes, or contact feasibility. For non-prehensile manipulation, force planning is also central because the object is not rigidly held. Even with a prescribed motion and contact mode, the contact force distribution determines whether the required wrench can be transmitted without slip or unloading \cite{mason2001mechanics,ruggiero2018nonprehensile}. However, less attention has been given to how large a friction margin should be maintained at each contact during execution, especially when the contacts are non-opposing and the required wrench changes over the motion. Bimanual tilting with an environmental pivot exposes this issue: losing either robot contact can remove the force needed to support or regulate the object.

\subsection{Contact Force Optimization and Friction Margins}
Optimization is widely used in contact-rich manipulation to reason about object motion, contact forces, and frictional constraints \cite{pang2023global,shirai2025hierarchical,zhang2025simultaneous}. For known contact geometry and motion, force planning becomes more structured and can be formulated as a convex optimization problem, such as a quadratic program or second-order cone program \cite{boyd2007fast,lippiello2012grasping,patankar2020hand}. Beyond computing feasible contact forces, an important question is how explicitly the optimized forces maintain margin from frictional limits. Friction margins have been used in broader manipulation optimization problems to improve feasibility and stability \cite{makita2013joint,shirai2024robust}. Such margins are useful, but they are not generally selected as local contact-wise force space margins. At the local contact level, conservatism can be introduced through the friction model or objective, such as using a reduced friction coefficient or adding penalty and barrier terms \cite{chen2021multi}. Cone shrinking provides a relative margin that vanishes as normal force approaches zero, while objective penalties depend on weights and compete with the force objective. Recent work has used a minimum friction margin to generate behavior that is distilled into a learning-based controller \cite{winkelbauer2024learning}. There, the margin serves as an intermediate data generation constraint rather than a constraint directly enforced during contact force planning, and its required value is fixed. When task force demand changes over a motion, a fixed margin can be insufficient in demanding configurations or add unnecessary loading when demand is small.

Together, these works leave open how to impose a contact-wise margin directly on planned contact forces, with a force space interpretation, while allowing the required margin to vary with task demand. We address this by enforcing a task-adaptive lower bound on the signed distance from each planned contact force to the nominal friction boundary as a hard constraint in a convex force planner.

\section{Method}\label{sec:method}
This section first introduces signed friction distance as a measure of force space margin. We then derive the pivot-centered task demand and the resulting affine task-adaptive margin for bimanual tilting, and embeds the margin in a convex force planner. Unless otherwise stated, vectors are expressed in the object frame and norms are Euclidean.

\subsection{Signed Friction Distance as a Force Space Margin}\label{sec:signed_friction_distance}
For a planar frictional point contact, let $\V n\in\R^2$ be the outward unit normal and let $\V f\in\R^2$ be the force applied to the object. We decompose $\V f$ into normal and tangential components as
\begin{equation}\label{eq:contact_force_decomposition}
    f_n=-\V n^\T\V f,
    \qquad
    \V f_t=\V f+f_n\V n ,
\end{equation}
where $f_n\geq0$ denotes compression. For a contact with friction coefficient $\mu>0$, the nominal Coulomb friction cone is
\begin{equation}\label{eq:nominal_friction_cone}
    \|\V f_t\| \leq \mu f_n,
    \qquad
    f_n\geq0 .
\end{equation}
Nominal feasibility only requires the force to lie inside this cone. It does not distinguish forces near the boundary from forces well inside the cone.

\begin{figure}[t] 
    \centering 
    \includegraphics[width=0.7\linewidth]{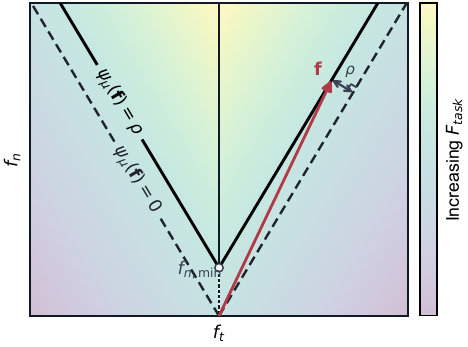} 
    \caption{Signed friction distance in the contact force plane. The nominal boundary $\psi_\mu(\V f)=0$ is the friction limit, while $\psi_\mu(\V f)\geq\rho$ keeps the planned force a finite signed distance inside the cone. When $f_t=0$, the same constraint induces a minimum normal force $f_{n,\min}$. The color bar schematically indicates increasing task force demand $F_\mathrm{task}$ and margin $\rho$.} 
    \label{fig:friction_margin_cone} 
\end{figure}
Figure~\ref{fig:friction_margin_cone} illustrates the margin measure used here: the signed distance from a contact force to the slip boundary in the local force plane. We define this quantity as \emph{signed friction distance},
\begin{equation}\label{eq:signed_friction_distance}
    \psi_{\mu}(\V f)
    =
    \frac{\mu f_n-\|\V f_t\|}{\sqrt{1+\mu^2}} .
\end{equation}
Thus, $\psi_{\mu}(\V f)>0$ inside the nominal cone, $\psi_{\mu}(\V f)=0$ on the slip boundary, and $\psi_{\mu}(\V f)<0$ outside the cone. It also relates the margin directly to force deviations. Let $\Delta\V f$ denote the force deviation expressed in the same nominal contact frame. Applying the decomposition in Eq.~\eqref{eq:contact_force_decomposition} to the perturbed force, the triangle inequality and Cauchy--Schwarz give
\begin{equation}\label{eq:signed_distance_degradation_bound}
    \psi_\mu(\V f+\Delta\V f)
    \geq
    \psi_\mu(\V f)-\|\Delta\V f\| .
\end{equation}
Therefore, let $\rho\geq0$ be the required margin, measured in force units. If the planned force satisfies
\begin{equation}
    \psi_\mu(\V f)\geq\rho ,
    \label{eq:friction_margin_constraint}
\end{equation}
then any realized force deviation with $\|\Delta\V f\|\leq\rho$ remains inside the nominal cone:
\begin{equation}
    \psi_\mu(\V f+\Delta\V f)
    \geq
    \psi_\mu(\V f)-\|\Delta\V f\|
    \geq
    \rho-\rho
    =
    0 .
\end{equation}
Thus, $\rho$ lower bounds the force space deviation that the planned contact force can tolerate before reaching the nominal slip boundary. The margin constraint in Eq.~\eqref{eq:friction_margin_constraint} is equivalent to the second-order cone (SOC) constraint,
\begin{equation}\label{eq:inner_friction_cone}
    \begin{bmatrix}
        \mu f_n - \rho\sqrt{1+\mu^2}\\
        \V f_t
    \end{bmatrix}
    \in
    \mathcal{L}^3,
\end{equation}
where $\mathcal{L}^3=\{(t,\V x)\in\mathbb R\times\mathbb R^2:t\geq\|\V x\|\}$ is the Lorentz cone. In particular, when $\V f_t=\V 0$, Eq.~\eqref{eq:inner_friction_cone} implies
\begin{equation}
    f_n
    \geq
    \frac{\rho\sqrt{1+\mu^2}}{\mu},
\end{equation}
so a positive margin also induces a minimum compressive normal force and discourages contact unloading. Moreover, the margin quantifies tolerance to friction coefficient overestimation. If $\mu$ is overestimated by $\delta\mu\geq0$, then, for $f_n>0$, the planned force remains feasible under the true coefficient if
\begin{equation}\label{eq:friction_coefficient_tolerance}
    \|\V f_t\|
    \leq
    (\mu-\delta\mu)f_n
    \quad\Longleftrightarrow\quad
    \delta\mu
    \leq
    \Delta\mu
    :=
    \frac{\mu f_n - \|\V f_t\|}{f_n},
\end{equation}
where $\Delta\mu$ is the exact overestimation tolerance. From Eq.~\eqref{eq:friction_margin_constraint},
\begin{equation}
    \mu f_n-\|\V f_t\|
    \geq
    \rho\sqrt{1+\mu^2}.
\end{equation}
Dividing by $f_n>0$, the imposed margin certifies friction feasibility for any overestimation satisfying
\begin{equation}
    \delta\mu
    \leq
    \frac{\rho\sqrt{1+\mu^2}}{f_n}.
\end{equation}
A conservatively reduced coefficient $\underline\mu$ can be used in Eq.~\eqref{eq:friction_margin_constraint} given a friction uncertainty bound, yielding $\psi_{\underline{\mu}}(\V f)\geq\rho$. Here, $\underline{\mu}$ explicitly encodes this bound, while \(\rho\) sets the minimum clearance from the reduced cone boundary in force units.

\subsection{Pivot-Centered Model for Bimanual Tilting}
\begin{figure}[t]
    \centering
    \includegraphics[width=0.7\linewidth]{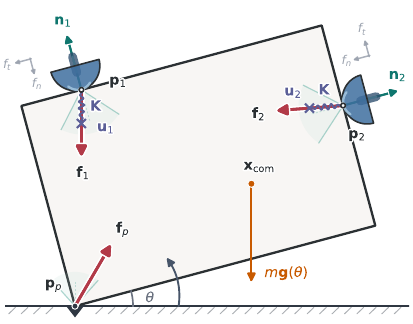}
    \caption{Bimanual tilting model. Two robot contacts apply forces to the object while an environmental pivot supplies a passive reaction.}
    \label{fig:task_model}
\end{figure}

\textbf{Problem formulation.} For the bimanual tilting task in Fig.~\ref{fig:task_model}, the object is rotated quasi-statically in $SE(2)$ about an environmental pivot by
two robot contacts. We assume that the (i) environmental contact occurs at a known object vertex, (ii) object motion is prescribed, and (iii) robot contacts remain sticking at known, fixed locations in the object frame.

\textbf{Notation.} The commanded tilt angle is denoted by $\theta$. Robot contacts are indexed by $i\in\{1,2\}$, and the pivot contact is indexed by $p$. Let $\V p_i$, $\V p_p$, and $\V x_\mathrm{com}\in\R^2$ denote the robot contact positions, pivot position, and nominal center of mass (CoM), respectively, all expressed in the object frame. For a planar lever arm $\V r=[r_x,r_y]^\T$, define $\V r^\perp=[-r_y,r_x]^\T$, so that the scalar moment generated by a force $\V f$ is $\V r\times\V f=(\V r^\perp)^\T\V f$.

\bigskip
For the tilting model in Fig.~\ref{fig:task_model}, the environmental pivot is passive, so its reaction force is not a decision variable. Given the robot contact forces, force balance determines
\begin{equation}\label{eq:pivot_force}
    \V f_p = -\V f_1 - \V f_2 - m\V g(\theta),
\end{equation}
where $m$ is the object mass and $\V g(\theta)$ is the gravity expressed in the object frame at tilt angle $\theta$. Thus, the planned force vector contains only the two robot contact forces,
\begin{equation}
    \V F = \begin{bmatrix}
                {\V f_1}^{\T} &
                {\V f_2}^{\T}
            \end{bmatrix}^{\T} \in \R^4 .
\end{equation}
Taking moments about the pivot eliminates the pivot reaction, since its moment arm is zero. The resulting task balance is
\begin{equation}\label{eq:moment_constraint}
    \V G \V F + \tau_g(\theta) = 0,
\end{equation}
where
\begin{equation}
    \V G = \begin{bmatrix}
                \V G_1 & \V G_2
            \end{bmatrix}, \quad
    \tau_g(\theta) = (\V x_\mathrm{com}-\V p_p)^{\perp\T} m\V g(\theta) \label{eq:tilting_gravity_moment}.
\end{equation}
Here, $\V G_i=(\V p_i-\V p_p)^{\perp\T}\in\R^{1\times2}$ maps the robot contact force $\V f_i$ to its moment contribution about the pivot, and $\tau_g(\theta)$ is the gravitational moment.

\subsection{Task-Adaptive Margin}\label{sec:tilting_margin_law}
The task force demand is the minimum-norm robot contact force vector needed to balance this pivot moment:
\begin{equation}\label{eq:task_force_component_opt}
    \V F_\mathrm{task}(\theta)
    =
    \arg\min_{\V F}
    \frac{1}{2}
    \left\|
        \V F
    \right\|^2
    \quad
    \mathrm{s.t.}
    \quad
    \V G\V F+\tau_g(\theta)=0 .
\end{equation}
Because the robot contacts must have nonzero moment-generating capability about the pivot, we assume $\V G\neq\V 0$. The minimum-norm task force in Eq.~\eqref{eq:task_force_component_opt} can equivalently be written as
\begin{equation}\label{eq:tilting_task_force_pseudoinverse}
    \V F_\mathrm{task}(\theta)
    =
    -\V G^\T
    \left(
        \V G\V G^\T
    \right)^{-1}
    \tau_g(\theta).
\end{equation}
The corresponding scalar demand is
\begin{equation}\label{eq:task_force_scalar}
    F_\mathrm{task}(\theta)
    =
    \left\|
        \V F_\mathrm{task}(\theta)
    \right\|
    =
    \left|
        \tau_g(\theta)
    \right|
    \left(
        \V G\V G^\T
    \right)^{-1/2}.
\end{equation}
This scalar demand measures the pivot moment demand after mapping it through the minimum-norm contact force map, before the margin or objective-dependent force redistribution.

The same minimum-norm map converts a scalar pivot-moment deviation $\delta\tau_g$ into a robot contact force correction of magnitude
\begin{equation}\label{eq:tilting_force_correction_norm}
    \left\|
        \Delta\V F
    \right\|
    =
    \left|
        \delta\tau_g
    \right|
    \left(
        \V G\V G^\T
    \right)^{-1/2}.
\end{equation}
This correction magnitude provides the task-level force deviation bound used to
allocate the robot contact margins. Consider deviations in object mass and CoM location. Perturbing Eq.~\eqref{eq:tilting_gravity_moment} to first order gives
\begin{equation}\label{eq:tilting_mass_com_split}
    \delta\tau_g
    =
    \frac{\delta m}{m}
    \tau_g(\theta)
    +
    m
    \left(
        \delta\V x_\mathrm{com}
    \right)^{\perp\T}
    \V g(\theta).
\end{equation}
If the property deviations are bounded as $\left|\delta m\right|\leq\Delta m$ and $\left\|\delta\V x_\mathrm{com}\right\|\leq\Delta x$, then
\begin{equation}\label{eq:tilting_moment_deviation_bound}
\begin{aligned}
    \left|
        \delta\tau_g
    \right|
    &\leq
    \frac{\Delta m}{m}
    \left|
        \tau_g(\theta)
    \right|
    +
    mg\Delta x .
\end{aligned}
\end{equation}
Combining Eqs.~\eqref{eq:tilting_force_correction_norm} and \eqref{eq:tilting_moment_deviation_bound} gives the sufficient task-level margin
\begin{equation}\label{eq:tilting_required_margin_bound}
    \rho_\mathrm{req}(\theta)
    =
    \left[
        \frac{\Delta m}{m}
        \left|
            \tau_g(\theta)
        \right|
        +
        mg\Delta x
    \right]
    \left(
        \V G\V G^\T
    \right)^{-1/2}.
\end{equation}
Using Eq.~\eqref{eq:task_force_scalar}, this motivates an affine task-adaptive margin
\begin{equation}\label{eq:tilting_affine_margin}
    \rho(\theta)
    =
    \lambda_m F_\mathrm{task}(\theta)
    +
    \lambda_\mathrm{CoM}.
\end{equation}
When such property deviation bounds are identified, a conservative parameter choice is
\begin{equation}\label{eq:tilting_margin_parameters}
\lambda_m
\geq
\frac{\Delta m}{m},
\qquad
\lambda_\mathrm{CoM}
\geq
mg\Delta x
\left(\V G\V G^\T\right)^{-1/2}.
\end{equation}
The contact-$i$ block of the minimum-norm correction allocates the margin to robot contact $i$ as
\begin{equation}\label{eq:contact_specific_margin}
    \rho_i(\theta)
    =
    \|\V G_i\|
    \left(\V G\V G^\T\right)^{-1/2}
    \rho(\theta).
\end{equation}
The preceding force correction derivation applies to the robot contacts and does not directly bound the passive pivot reaction. We therefore heuristically define the pivot margin as
\begin{equation}\label{eq:pivot_margin}
    \rho_p(\theta)
    =
    \rho_1(\theta)+\rho_2(\theta).
\end{equation}

The model-based force deviation derivation determines the configuration dependent structure of task-adaptive margin in Eq.~\eqref{eq:tilting_affine_margin}.

\subsection{Contact Force Planning}
For each tilt angle $\theta$, the optimizer computes robot contact forces that satisfy pivot moment balance and signed friction distance margin constraints:
\begin{equation}\label{eq:force_planning_objective}
\begin{aligned}
    \V F^*(\theta) =
    \arg\min_{\V F \in \R^4} \quad&
    \frac{1}{2}
    \sum_{i=1}^{2}
    \|\V f_i\|^2 \\
    \mathrm{s.t.}\quad&
    \V G \V F + \tau_g(\theta) = 0,\\
    &
    \V f_p = -\V f_1-\V f_2-m\V g(\theta),\\
    &
    \psi_{\mu_i}(\V f_i) \geq \rho_i(\theta),
    \qquad i\in\{1,2\},\\
    &
    \psi_{\mu_p}(\V f_p) \geq \rho_p(\theta).
\end{aligned}
\end{equation}
The contact-specific margins $\rho_i(\theta)$ are applied to the robot contacts, while $\rho_p(\theta)$ is applied to the pivot reaction, with each contact using its corresponding friction coefficient.

For fixed $\theta$, $\rho_i(\theta)$ and $\rho_p(\theta)$ are fixed parameters. The moment constraint is linear in $\V F$, and each friction margin constraint has the SOC form in Eq.~\eqref{eq:inner_friction_cone}. Therefore, Eq.~\eqref{eq:force_planning_objective} is an SOCP and can be solved using interior-point methods~\cite{lobo1998applications}. The task-adaptive margins only shift the lower bounds inside the same friction constraints, preserving convexity.

\subsection{Robot Execution}\label{sec:contact_control}
The optimized robot contact forces are executed through impedance control by converting each desired force into an equilibrium point. Let $\V q$ denote the object pose used for frame transformations, with tilde quantities expressed in robot $i$'s base frame. For each robot contact $i\in\{1,2\}$, the commanded Cartesian equilibrium point is
\begin{equation}\label{eq:u_contact}
    \tilde{\V u}_{i}^*(\V q)
    =
    \tilde{\V p}_i(\V q)
    +
    \V K_i^{-1}\tilde{\V f}_i^*(\theta).
\end{equation}
Here, $\V K_i\in\R^{2\times 2}$ is the robot translational stiffness matrix. The end-effector orientation follows the object tilt with a fixed object frame offset.

\section{Experiments}\label{sec:experiments}

\subsection{Setup}
\begin{figure}[t]
    \centering
    \includegraphics[width=\linewidth]{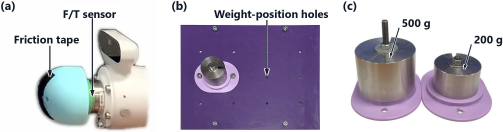}
    \caption{Experimental setup. (a) Kinova Gen3 robot arm with an ATI Mini40 force/torque sensor at the wrist and a hemispherical end-effector covered with friction tape. (b) Hole pattern plate for varying the object CoM. (c) Calibration weights used for varying the object mass.}
    \label{fig:exp_setup}
\end{figure}
As illustrated in Fig.~\ref{fig:overview}, the experiments used two 7-DoF Kinova Gen3 manipulators for bimanual tilting. Each wrist had an ATI Mini40 force/torque sensor and a hemispherical end-effector covered with friction tape, as shown in Fig.~\ref{fig:exp_setup}a. The manipulated box measured $0.3 \times 0.2 \times 0.2~\mathrm{m}$ and had a base mass of $1.7~\mathrm{kg}$. The nominal condition added a centered $200~\mathrm{g}$ calibration weight using the plate in Fig.~\ref{fig:exp_setup}b, giving a total mass of $1.9~\mathrm{kg}$ and defining the centered CoM model. The measured friction coefficients, averaged over three trials, were $\mu_1=\mu_2=0.75$ for the robot--object contacts and $\mu_p=0.5$ for the object--environment contact. The margin values $\lambda_m=0.25$ and $\lambda_\mathrm{CoM}=0.5~\mathrm{N}$ are design parameters selected once from the nominal setup and kept fixed across all conditions. Thus, the experiments evaluate the resulting contact behavior rather than identifying tight physical deviation bounds. For the fixed experimental contact geometry, the resulting margins were $\rho_1(\theta)\approx0.56\,\rho(\theta)$, $\rho_2(\theta)\approx0.83\,\rho(\theta)$, and $\rho_p(\theta)\approx1.39\,\rho(\theta)$. The optimization was implemented in CVXPY~\cite{diamond2016cvxpy} and solved with ECOS~\cite{domahidi2013ecos} at $0.1^\circ$ angle increments. Execution used impedance control with translational stiffness $1000~\mathrm{N/m}$ and rotational stiffness $20~\mathrm{Nm/rad}$.

\subsection{Evaluation Conditions}\label{sec:evaluation_conditions}
\begin{table}[t]
    \centering
    \caption{Object property and margin evaluation conditions. Mass and CoM offset define the optimizer model. Object property variations retain the same contact-specific margin law and parameters, while the no-margin condition uses nominal properties with $\rho_i=0$. CoM offsets are relative to the nominal.}
    \label{tab:evaluation_conditions}
    \footnotesize
    \resizebox{\linewidth}{!}{%
    \begin{tabular}{lcccc}
        \hline
        Condition & Varied factor & Mass (kg) & CoM offset (mm) & Margins \\
        \hline
        Nominal & -- & 1.9 & 0 & $\rho_i(\theta)$ \\
        Light object & Mass & 1.7 & 0 & $\rho_i(\theta)$ \\
        Heavy object & Mass & 2.2 & 0 & $\rho_i(\theta)$ \\
        Toward pivot CoM & CoM & 1.9 & $-6.3$ & $\rho_i(\theta)$ \\
        Away from pivot CoM & CoM & 1.9 & $+6.3$ & $\rho_i(\theta)$ \\
        No margin & Margin & 1.9 & 0 & $0$ \\
        \hline
    \end{tabular}
    }
\end{table}

\textit{Object property and margin evaluation.} Table~\ref{tab:evaluation_conditions} summarizes the evaluation conditions. The light object condition removes the calibration weight, while the heavy object condition replaces it with a centered $500~\mathrm{g}$ weight. The CoM variation conditions place the $200$ g weight at plate holes displaced by $60~\mathrm{mm}$ from the centered position, producing the net CoM offsets listed in Table~\ref{tab:evaluation_conditions}. The no margin ablation uses the nominal object properties with $\rho_i=0$.

\textit{Model mismatch evaluation.} With the physical setup fixed at the nominal condition, we independently evaluated three model mismatches. The optimizer assumed a mass \(10\%\) below nominal, a CoM shifted toward the pivot by \(10\%\) of the object half-width, and friction coefficients \(10\%\) above their measured values. All other experimental settings were unchanged.

For all conditions, five trials were conducted with the object tilted leftward. Success required completing the tilt while maintaining both robot contacts. Failure corresponded to severe slip, contact loss, or object drop preventing completion.

\section{Results}\label{sec:results}
In this section, we evaluate the proposed method from five perspectives:
\begin{enumerate}[leftmargin=*]
    \item whether the optimized contact forces follow the task mechanics under nominal tilting,
    \item whether the task-adaptive margin captures object property effects through the task-demand model,
    \item whether the method remains effective under model mismatch,
    \item whether task-adaptive margins improve robustness over a fixed margin,
    \item whether positive margins support successful task execution.
\end{enumerate}

\subsection{Contact Force Profiles and Margin Effects}
\begin{figure}[t]
    \centering
    \includegraphics[width=\linewidth]{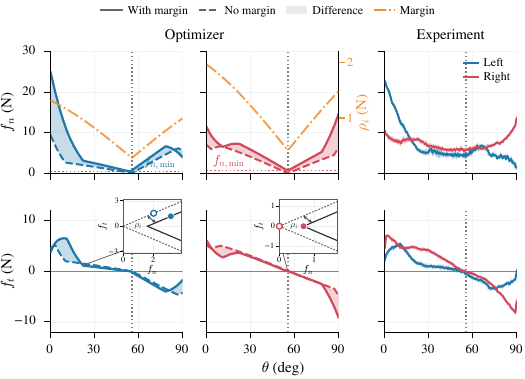}
    \caption{Nominal contact force profiles and margin effects. Optimized columns compare robot contact forces with and without margins. Shading indicates their difference. Orange curves show contact-specific margins $\rho_i(\theta)$. Dotted lines mark $f_{n,\min}$ and $\theta_{\mathrm{tip}}$. Insets show force vectors, with open and filled markers denoting no margin and margin solutions. The experiment column overlays all trials and their means.}
    \label{fig:nominal_margin_profile_with_exp}
\end{figure}

We first examine how the optimized forces follow the task mechanics in the nominal condition and how the margin changes their distance from the friction boundary. Figure~\ref{fig:nominal_margin_profile_with_exp} compares optimized force profiles with and without the task-adaptive margin, together with experimental force profiles.
\begin{figure*}[t]
    \centering
    \includegraphics[width=0.95\linewidth]{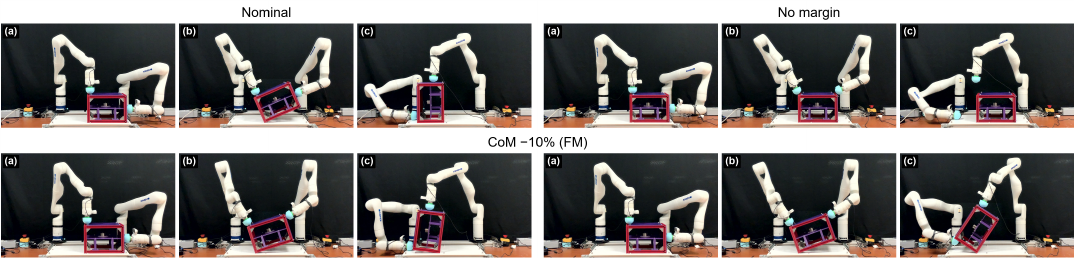}
    \caption{Representative tilting sequences under nominal, no margin, and fixed margin CoM mismatch conditions. The nominal trial completes the tilt with both contacts maintained, while removing the margin leads to contact loss. Both fixed margin trials exhibit severe slip: the bottom left terminates before completing the tilt, and the bottom right loses the left contact as it slips below the box.}
    \label{fig:tilting_sequence}
\end{figure*}

For the box with centered CoM, the geometric tipping line occurs at approximately $\theta_{\mathrm{tip}}=55^\circ$, where the gravity line of action nearly aligns with the environmental pivot. The gravitational moment $\tau_g(\theta)$ about the pivot and the optimized force demand therefore decrease near $\theta_{\mathrm{tip}}$ and increase after the object passes this configuration. The optimized tangential forces $f_t$ also transition from support to braking. Under the contact force sign convention in Fig.~\ref{fig:task_model}, the tangential forces are positive before $\theta_{\mathrm{tip}}$, when the robot contacts oppose gravity to support rotation, and negative afterward, when gravity assists the rotation. The experimental responses follow the optimized profiles and show the same transition, although a magnitude offset remains, consistent with pose deviations under impedance execution and unmodeled contact effects.

Figure~\ref{fig:nominal_margin_profile_with_exp} also shows how the margins affect the optimized force distribution. Both margins decrease near $\theta_{\mathrm{tip}}$ as task force demand falls, while the right contact margin remains larger because that contact has greater moment leverage about the pivot, as reflected in Eq.~\eqref{eq:contact_specific_margin}. Near tipping, the normal forces remain above the no margin solution due to the force-independent term, preventing the optimized normal forces from collapsing toward zero. The force plane insets show these effects geometrically: the margin moves the optimized force away from the nominal friction boundary at early tilt, and retains a positive normal force near tipping. The tangential forces also change to satisfy the same moment balance with the full contact force vectors.

We also measured solve time at each tilt angle. The no margin case sets $\rho_i(\theta)=0$, while the with margin case changes only the lower bound in the same SOC friction constraint. Both preserve the same SOCP structure. Mean solve times were \(8.99\pm1.22\) ms with the margin and \(9.05\pm1.45\) ms without it, indicating comparable computational costs.

\subsection{Adaptation to Object Property Variations}
\begin{figure}[t]
    \centering
    \includegraphics[width=\linewidth]{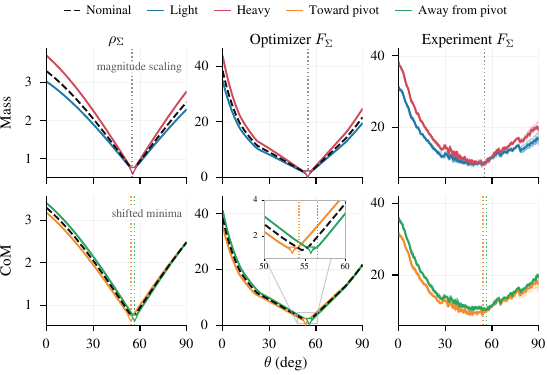}
    \caption{Adaptation to object property variations. Rows show mass and CoM variations. Columns show $\rho_\Sigma$, optimized and experimental $F_\Sigma$, all in newtons. Mass changes scale the profiles, CoM changes shift their minima. Markers indicate profile minima. Vertical dotted lines mark $\theta_{\mathrm{tip}}$.}
    \label{fig:task_adaptive_margin_force}
\end{figure}

We next examine how the margin adapts to object property changes through the task-demand model and how this adaptation affects the optimized force profiles. Figure~\ref{fig:task_adaptive_margin_force} shows the summed margin $\rho_\Sigma=\rho_1+\rho_2$ and combined force magnitude $F_\Sigma=\|\V f_1\|+\|\V f_2\|$ for the variations listed in Table~\ref{tab:evaluation_conditions}.

In the mass sweep, changing mass scales the gravitational moment over the motion without changing the tipping angle. Increasing mass produces larger $\rho_\Sigma$ and optimized $F_\Sigma$, while the lighter object produces lower profiles. This illustrates the role of the task-dependent gain: scaling the margin with $F_\mathrm{task}(\theta)$ places a larger margin at high demand while reducing the optimized loading near tipping configuration.

In the CoM sweep, shifting the CoM changes the moment arm in $\tau_g(\theta)$. The resulting moment differences are larger at early tilt angles, where the gravity lines of action have more distinct perpendicular distances to the pivot. These distances converge toward the final configuration, bringing the moment demands, margins, and force profiles closer together. The CoM shift also changes where the gravity line passes through the pivot, shifting the profile minima. This sensitivity motivates the force-independent margin term: CoM deviations can induce nonzero force corrections even when at low nominal task demand, thus this term prevents the margin from collapsing in these configurations.

\begin{table}[t]
    \centering
    \caption{Real-world experiment outcomes. Successful trials complete the tilt with both robot contacts maintained. Force standard deviation summarizes trial-to-trial variation in experimental contact forces and is omitted for failed conditions. Violation rate is the percentage of samples with $f_n\leq0.5~\mathrm{N}$ or $|f_t|/(\mu f_n)>1$. $^{\dagger}$Omitted as failed trials were terminated early for safety. FM: fixed margin.}
    \label{tab:combined_task_ablation_results}
    \footnotesize
    \resizebox{\linewidth}{!}{%
    \begin{tabular}{lccc}
        \hline
        Condition & Success trials & Force std. dev. (N) & Violation rate (\%) \\
        \hline
        Nominal & 5/5 & 0.46 & 6.9 \\
        Light object & 5/5 & 0.50 & 3.9 \\
        Heavy object & 5/5 & 0.40 & 4.0 \\
        Toward pivot CoM & 5/5 & 0.38 & 3.3 \\
        Away from pivot CoM & 5/5 & 0.36 & 2.9 \\
        No margin & 0/5 & -- & 90.7 \\
        \noalign{\vskip 1pt}
        \multispan{4}\hskip 6pt\leaders\hrule height 0.2pt\hfill\hskip 6pt\cr
        \noalign{\vskip 1pt}
        Mass $-10\%$ & 5/5 & 0.59 & 9.1 \\
        CoM $-10\%$ & 5/5 & 0.63 & 7.9 \\
        Friction $+10\%$ & 5/5 & 0.64 & 7.8 \\
        \noalign{\vskip 1pt}
        \multispan{4}\hskip 6pt\leaders\hrule height 0.2pt\hfill\hskip 6pt\cr
        \noalign{\vskip 1pt}
        Nominal (FM) & 5/5 & 1.12 & 7.5 \\
        CoM $-10\%$ (FM) & 0/5 & -- & --$^{\dagger}$ \\
        \hline
    \end{tabular}
    }
\end{table}

Table~\ref{tab:combined_task_ablation_results} summarizes the execution outcomes. All trials under object property variations completed with contact maintained. These results support the adaptability of the margin law: using the same parameters, the margin follows $F_\mathrm{task}(\theta)$ under mass and CoM variations, producing the expected scaling and shifting trends while maintaining contact across all tested conditions.

\subsection{Robustness to Model Mismatch}
Under the model mismatch conditions defined in Section~\ref{sec:evaluation_conditions}, all trials completed successfully, although their violation rates exceeded those in the model matched trials, shown in Table~\ref{tab:combined_task_ablation_results}. This increase is expected because inaccurate model parameters can misrepresent the actual force requirements or contact feasibility. Specifically, underestimating mass reduces the predicted gravitational moment, while shifting the assumed CoM toward the pivot reduces it at early tilt angles. Both can underestimate the required forces. Nevertheless, the margin constraints keep the optimized forces away from the friction boundary, supporting successful task completion despite these discrepancies.

\begin{figure}[t]
    \centering
    \includegraphics[width=\linewidth]{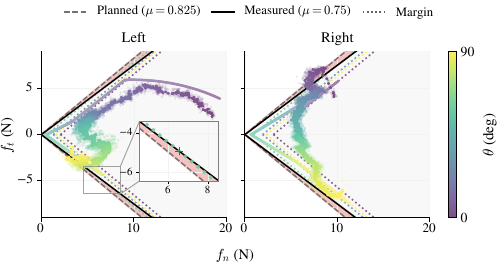}
    \caption{Friction mismatch at robot contacts. Dashed and solid lines denote the planning and measured friction cones, respectively. Red shading marks the region feasible only under the planning coefficient. Dotted lines show margin boundaries at $30^\circ$ intervals. Optimized forces and experimental samples are colored by tilt angle. Inset magnifies one boundary's intersection with the measured cone.}
    \label{fig:friction_mismatch}
\end{figure}
The margins also provide tolerance to friction coefficient overestimation, with the admissible error depends on both the imposed margin and the optimized force, as quantified by Eq.~\eqref{eq:friction_coefficient_tolerance}. Figure~\ref{fig:friction_mismatch} shows that overestimation enlarges the planning cone, allowing portions of the margin boundary to lie outside the measured cone. The inset magnifies one such intersection. Increasing the margin shifts the boundary inward and moves its intersection with the measured cone farther from the cone apex, providing greater tolerance to friction uncertainty. Under the tested overestimation, the optimized forces remained inside the measured cone, and all trials completed successfully. Together, these results support the role of margins in tolerating model uncertainty during real-world execution.

\subsection{Fixed Margin Study}
The task-adaptive margin follows from the model-derived force correction required to accommodate uncertainty and therefore varies with task demand. To examine the role of this configuration dependence, we compare it with a fixed margin under model-matched and mismatched conditions. For each condition, the constant margin \(\rho_{\mathrm{fix}}\) is calibrated to match the task-adaptive solution's average combined force magnitude, $\overline{F_\Sigma} =\frac{1}{N}\sum_{n=1}^{N} \left(\|\V f_1(\theta_n)\|+\|\V f_2(\theta_n)\|\right)$. Within each condition, only the margin schedule differs.

Table~\ref{tab:combined_task_ablation_results} shows that both formulations completed all nominal trials. Under CoM mismatch, however, the task-adaptive formulation completed all trials, whereas the fixed margin formulation failed all. The bottom sequences in Fig.~\ref{fig:tilting_sequence} illustrate two such failures. Both show severe slip: the left sequence terminates before completing the tilt, while the right sequence subsequently loses the left contact as it slips below the box. Neither satisfies the success criterion in Section~\ref{sec:evaluation_conditions}.

\begin{figure}[t]
    \centering
    \includegraphics[width=\linewidth]{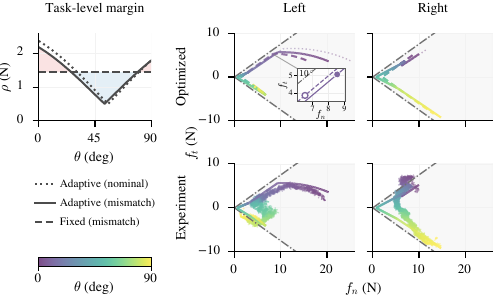}
    \caption{Task-adaptive and fixed margins under CoM mismatch with matched average combined force magnitude. Left: task-level margins, with red (blue) shading where the fixed margin is lower (higher) than the task-adaptive margin. Force plane panels show optimized profiles (top) and task-adaptive experimental samples (bottom) for both contacts. Inset magnifies the optimized forces at $\theta=10^\circ$. Line style denotes formulation and color denotes tilt angle.}
    \label{fig:fixed_margin_mismatch}
\end{figure}
Figure~\ref{fig:fixed_margin_mismatch} relates these outcomes to how margin is distributed across configurations. The top-left panel shows that the fixed margin is smaller than the task-adaptive margin during the higher demand portions at the beginning and end of the tilt, but larger near minimum task demand, as indicated by the red and blue shading. Thus, matching average combined force magnitude does not reproduce the configuration dependent allocation of protection. Through the coupled equilibrium constraints, this redistribution changes both normal and tangential forces, as shown in the top force plane panels. At $\theta=10^\circ$, the inset shows that the fixed margin solution produces lower $f_n$ and $f_t$ and lies closer to the friction boundary. Under model mismatch, this provides less local tolerance for underestimated task demand, consistent with the observed failures. Together, these outcomes support allocating margin according to task demand rather than using a constant value.

\subsection{No Margin Ablation Study}
\begin{figure}[t]
    \centering
    \includegraphics[width=\linewidth]{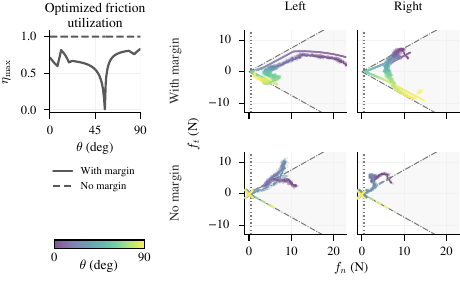}
    \caption{Left: optimized friction utilization $\eta_{\max}$ with and without margins. Right: experimental robot contact forces over trials for the with margin (top row) and no margin (bottom row) conditions. Points show experimental forces, curves show optimized forces, with color indicates tilt angle. Shading shows the robot--object friction cone, and crosses mark $f_n\leq0.5~\mathrm{N}$, indicating contact loss.}
    \label{fig:margin_ablation}
\end{figure}

We use the no margin ablation to test whether nominal friction feasibility alone is sufficient for maintaining contact. In this ablation, $\rho_i(\theta)=0$ at all contacts while retaining the nominal object properties. The top-left panel of Fig.~\ref{fig:margin_ablation} reports the optimized robot contact friction utilization, $\eta_{\max}(\theta) =
    \max_{i\in\{1,2\}}
    \frac{|f_t^{(i)}(\theta)|}{\mu f_n^{(i)}(\theta)}$, 
where values near one indicate forces close to the nominal friction boundary. The no margin solution stays near $\eta_{\max}=1$, indicating little distance from the boundary. In contrast, the with margin solution maintains lower utilization by keeping the optimized forces away from the boundary.

The right panels of Fig.~\ref{fig:margin_ablation} show the experimental contact forces across all trials. The with margin samples remain mostly inside or near the friction cone and above the normal force threshold, whereas the no margin samples frequently lie outside the cone or below the threshold. This contrast is consistent with the severe slip and contact loss observed in Fig.~\ref{fig:tilting_sequence}. We count a measured sample as a violation if $f_n\leq0.5~\mathrm{N}$, or if $|f_t|/(\mu f_n)>1$ for samples with $f_n>0.5~\mathrm{N}$, and average this fraction over contacts and trials. Table~\ref{tab:combined_task_ablation_results} shows that the with margin nominal condition completed all trials with a $6.9\%$ violation rate, whereas the no margin ablation failed all trials and reached $90.7\%$. These results indicate that, in the tested setup, enforcing a finite lower bound on signed friction distance helps maintain contact compared with nominal friction feasibility alone.

\section{Conclusion}\label{sec:conclusion}
This work presented signed friction distance margins for contact force planning in bimanual non-prehensile manipulation. Lower bounding each planned force's distance to the nominal friction boundary yields a force space margin directly comparable to force deviations. For bimanual tilting, pivot moment deviations yield an affine task-adaptive margin, allocated to robot contacts by their moment contributions and combined with a heuristic pivot margin in a convex planner. The optimized and measured force profiles reflected the evaluated object property variations, and all task-adaptive trials completed successfully under all tested task-adaptive conditions. At matched average combined force magnitude, the fixed margin completed the nominal trials but failed under model mismatch, supporting configuration dependent margin allocation to accommodate model uncertainty. The no-margin ablation produced a larger violation fraction that led to contact loss and task failure.

The same margin principle can be generalized by defining the task-level margin from the largest minimum-norm contact force correction over admissible wrench deviations, with parameters set by the deviation representation. They add no decision variables to a fixed mode solve. For problems with multiple candidate modes, a higher-level search could prune modes lacking sufficient margin before sequence planning, while closed-loop extensions could update margins, forces, and modes online from state estimates. Overall, these results support lower bounding signed friction distance as a physically interpretable means of moving contact force planning beyond nominal friction feasibility toward force space margins.

\appendix[General Task-Adaptive Margin]\label{app:general_margin}
For a general quasi-static task at configuration $\V q$, consider $\V G(\V q)\V F+\V w_{\mathrm{ext}}(\V q)=\V 0$, where the external wrench may be multidimensional. For an admissible wrench deviation set $\mathcal D(\V q)$, define the minimum-norm contact force correction as
\begin{equation*}
\begin{aligned}
    \Delta\V F^*(\V q,\delta\V w)
    &=\arg\min_{\Delta\V F}\frac{1}{2}\|\Delta\V F\|^2\\
    &\quad\mathrm{s.t.}\quad\V G(\V q)\Delta\V F+\delta\V w=\V 0.
\end{aligned}
\end{equation*}
Assuming all admissible deviations admit balancing corrections and the maximum exists, the general task-level margin is
\begin{equation*}
    \rho(\V q)=\max_{\delta\V w\in\mathcal D(\V q)}
    \|\Delta\V F^*(\V q,\delta\V w)\|.
\end{equation*}

\balance

\printbibliography

\end{document}